\documentclass[journal]{IEEEtran}  
\usepackage{caption}

\usepackage{float}
\usepackage{cite}                  
\usepackage{amsmath, amssymb,amsfonts}     
\usepackage[table, svgnames, dvipsnames]{xcolor}  
\definecolor{LightGray}{gray}{0.9}                

\usepackage{graphicx}
\usepackage{subcaption}
\usepackage{algorithm}
\usepackage{algorithmic}
\usepackage{booktabs}
\usepackage{multirow}
\usepackage{url}
\usepackage{hyperref}
\usepackage{soul}     
\sethlcolor{gray}
\usepackage{xspace}
\usepackage{orcidlink}
\usepackage{slashed}

\usepackage{balance}

\begin{document}

\title{Detection of Tampering in Wireless Electrocardiogram Using Hybrid Machine Learning}
\author{Siddhant Deshpande,Yalemzerf Getnet~\orcidlink{0009-0008-5975-2086}, Waltenegus Dargie~\orcidlink{0000-0002-7911-8081}, \IEEEmembership{Senior Member, IEEE}  
    \thanks{Manuscript submitted on 04 August 2025.}
    \thanks{S. Deshpande and W. Dargie are with the Faculty of Computer Science, Technische Universit{\"a}t Dresden, 01062 Dresden, Germany (e-mail: siddhantdeshpande3@gmail.com, waltenegus.dargie@tu-dresden.de)}
    \thanks{Y. Getnet is with the Department of Electrical and Computer Engineering at Addis Ababa University, Ethiopia (e-mail:  yalemzerf.getnet@aau.edu.et)}}
\maketitle

\begin{abstract}
With the proliferation of wireless electrocardiogram (ECG) systems for health monitoring and authentication, protecting signal integrity against tampering is becoming increasingly important. This paper analyzes the performance of CNN, ResNet, and hybrid (Transformer+CNN) models for tamper detection. Six tampering strategies, including structured segment substitutions and random insertions, are emulated to simulate real-world attacks. The one-dimensional ECG signals are converted into a two-dimensional representation in the time-frequency domain using the continuous wavelet transform (CWT). The models are trained and evaluated using ECG data from 54 individuals recorded in four separate batches outside of clinical settings while the individuals performed seven different daily activities. Experimental results show that the hybrid models achieved an accuracy of over 99.5\% in highly fragmented tampering scenarios. Even with subtle manipulations (e.g., 50\%–50\% and 75\%–25\% substitutions), the hybrid models demonstrated consistently reliable performance, achieving an average accuracy of 98\%. However, the hybrid models were relatively computationally expensive.
\end{abstract}


\section{Introduction}
\label{sec: introduction} 

The electrocardiogram (ECG) plays an important role in the diagnosis and monitoring of cardiovascular diseases \cite{Sumalatha2024Deep,Abubaker2023Detection}. Traditionally, measurements are performed in controlled environments by medically trained personnel. This implicitly guarantees their quality and authenticity. With growing interest in preventive medicine and long-term monitoring of cardiac patients, wireless ECGs are increasingly being used outside of clinical settings \cite{Behfar2021Fully,kassem2021context}. This practice is gaining ground because it enables cost-effective and scalable monitoring, but it also raises concerns about quality and authenticity \cite{Uwaechia2021Comprehensive,Ribeiro2018Evolution}. For the measurements to be clinically relevant, these concerns must be addressed. The goal of this article is to address the issue of ECG manipulation.

\begin{figure}
  \centering
\includegraphics[width=0.5\textwidth]{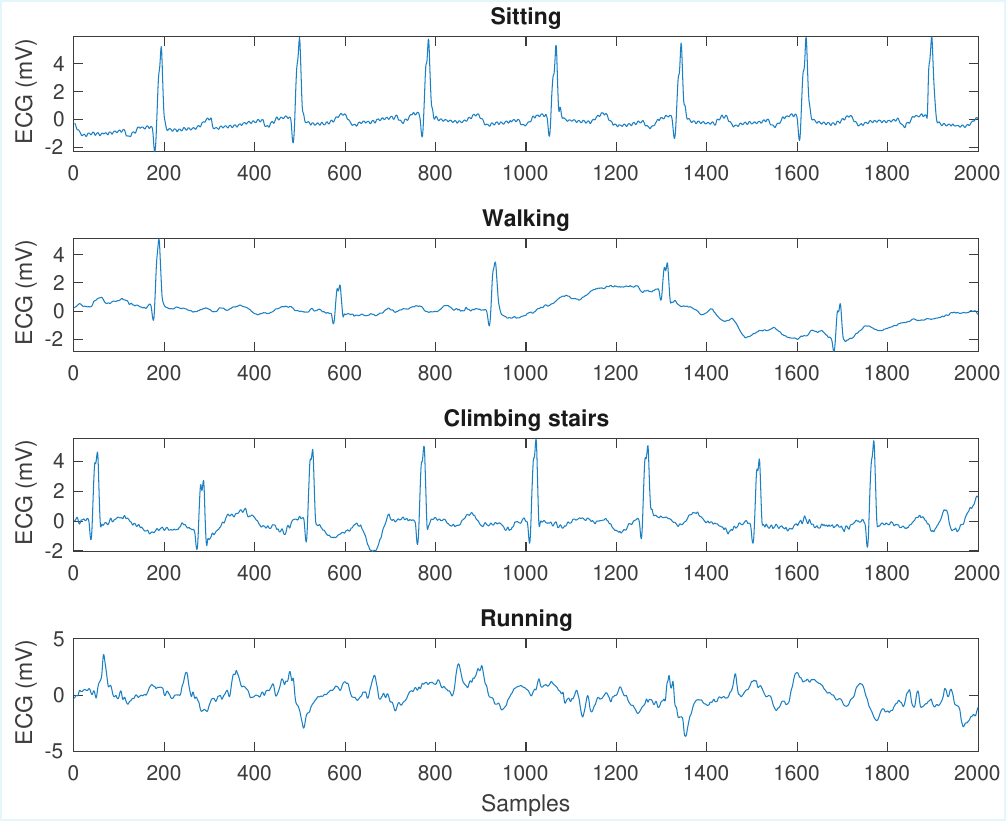}%
  \caption{Snapshots of electrocardiogram measurements taken when a female subject carried out four different activities (sitting, walking, climbing stairs, and running).}
  \label{fig:ecg}
\end{figure}

In a clinical context, the ECG measurement is a legal document \cite{sassi2017pdf, wagner2014marriott}. Decisions made based on this document are binding and can have serious consequences \cite{kligfield2007recommendations}. When electrocardiograms are used outside of clinical settings, their authenticity cannot be guaranteed and they can be compromised in various ways. One of the most obvious dangers is the classic man-in-the-middle problem \cite{eberz2017broken}. However, measurements can be intentionally altered by the wearer of the device for various reasons, such as to conceal or mimic certain heart conditions, to mislead employers, family members, or the legal system, or to make illegible claims against insurance policies\cite{Wu2021,awadallah2024artificial}. The problem is exacerbated by the fact that heart rhythms and ECG morphology change significantly when subjects move freely or exercise, making distinguishing spurious from authentic components a formidable challenge\cite{shandhi2020estimation,ji2024ecg}. This is evident in Fig.~\ref{fig:ecg}, which shows snapshots of a subject's ECG while performing various exercises on the same day.

In this paper, we emulate various fine-grained (less than one second in scope) manipulation strategies and propose different machine learning models—CNN-based, ResNet-based, Transformer-based, and hybrid (Transformer+CNNs)—for their detection. The models exploit correlations between salient ECG features to capture unique (for an individual) and persistent (across different activities) features in ECG measurements during highly fluctuating cardiac workloads. We train and test these models on 54 individuals performing seven different physical activities. Nineteen of our individuals were female, the remainder male. Twenty of our individuals lived in India, the remainder in Germany. The results show that the ResNet and the hybrid CNN-Transformer model achieved high accuracy in manipulation detection. The accuracy of the CNN and ResNet models in detecting highly localized manipulations is high; however, they were sensitive to variations in hyperparameters. The hybrid Transformer-CNN-based models achieved high accuracy, as well as a high degree of stability and consistent training results. However, we will show that this was associated with high computational complexity. The contributions of this article can be summarized as follows:

\begin{itemize}
\item \textbf{Data:} Our approach is real-world data-driven. A large volume of ECG data was collected from 54 subjects carrying out 7 different activities. The data were collected in four separate batches between 2019 and 2025 and exhibit a high variance, both across subjects and across activities. 
\item \textbf{Hybrid model for tampering detection:} Multiple hybrid machine learning frameworks are  proposed for tampering detection and the models are evaluated against different fine-grained  tampering strategies. The hybrid models aim to capture correlations at different abstraction levels. \item\textbf{Experiments:} Repeated experiments were conducted to assess the performance and stability of the models in the presence of highly variable inputs. Overall, the hybrid models achieve a detection average accuracy exceeding 98\%.
\end{itemize}
The remainder of this paper is organized as follows. Section~\ref{sec:related} provides a review of related work in the field. Section~\ref{sec:data} describes the data acquisition process. In Section~\ref{sec:features}, we outline the strategies used for feature identification and extraction. Section~\ref{sec:model} details the proposed model architecture, including parameter selection and configuration. Section~\ref{sec:evaluation} presents the evaluation methodology along with quantitative results. Finally, Section~\ref{sec:conclusions}, offers concluding remarks and discusses potential directions for future research.

\section{Related Work}
\label{sec:related} The widespread use of wireless ECG data for healthcare monitoring and person identity verification has led to concern about the wireless ECG data's authenticity\cite{tasnim2023cardiovascular,9427539}. Various ECG data tampering detection mechanisms  are proposed  to improve the reliability of wireless ECG data, and the proposed approaches are discussed in this section.

In \cite{24}, the authors proposed a Quantum Arrhythmia Detection System (QADS). The system developed secure ECG data by using quantum blockchain technology, which uses a blockchain algorithm that integrates a controlled quantum walk hash function and a quantum authentication protocol during block creation. To detect irregular heartbeats,  temporal features are extracted from securely stored ECG data  by using a hybrid quantum convolutional neural network (HQCNN). The method is tested on the MIT-BIH Arrhythmia database, and  HQCNN achieved an accuracy of 94.3\% and CNN achieved an accuracy of 92.5\%.

Manipulatable Haar Transform (MHT) technique,  a non-invertible transformation technique, is developed in \cite{25}. The method is used to identify individuals based on ECG data,  and it is also used  to protect wireless ECG data against potential threats. 
A stable temporal feature vector is first extracted from the ECG signal
collected. Then, the MHT is performed  on the extracted ECG feature vector to convert it to another version in a noninvertible manner. This conceals the sensitive information contained in the extracted feature vector. A second feature vector extracted
synchronically from the same user by another sensor node is transformed in the same way as the first feature vector. Finally, authentication takes place in the encrypted domain by comparing the transformed versions of the two feature vectors.
The simulation result tested on PTB dataset show that an equal error rate (EER) 7.62\% is achieved.

A study in \cite{26} proposed dual-attention W-Net (a DAW-Net), a dual-attention-based dual encoder-decoder architecture designed to separate maternal (mECG) and fetal (fECG) signals from abdominal ECG (aECG) signals affected by noise from diverse sources. The network extracts feature maps related to both maternal and fetal QRS complexes, enabling separation of mECG and fECG signals. Correlation attention enhances fetal QRS by masking relevant regions, while self-attention captures maternal QRS context for removal, and skip connections amplify QRS signal clarity. The model was evaluated using two real datasets, FECGSYNDB\footnote{https://physionet.org/content/fecgsyndb/1.0.0/sub01/} and ABFECGDB\footnote{https://www.physionet.org/content/adfecgdb/1.0.0/}, and  obtained F1 score of 98.13\%. 

CNN and self-supervised contrastive learning were developed in \cite{27} to extract user-specific features from unlabeled ECG data. CNN was trained on large, unlabeled datasets by using contrastive learning, and this learning  method is used to improve model generalization. The model is tested using MIT-BIH and ECG-ID datasets and achieved 99.15\% accuracy. 

The study in \cite{28} proposed the Statistical N-best Adaptive Fourier Decomposition (SAFD), a method that  extends the traditional n-best Adaptive Fourier Decomposition (AFD) \cite{28b} from single-signal to multi-signal processing and operates within the stochastic Hardy space. The SAFD method learns some atoms that effectively capture the internal structure of ECG signals  and can produce a discriminative representation that preserves their time-frequency characteristics.
The model is used for biometric identification, and it is evaluated on five ECG databases (Fantasia, MITDB, ECG-ID, EDB, and AED) to achieve an accuracy of 97.59\%.

The Singular Value Decomposition (SVD) method was proposed in \cite{29},  and the proposed method was applied to denoise the  ECG signals by extracting orthonormal eigenvectors that capture essential features and separate orthogonal components. 
 To validate the method,  a study was conducted involving human subjects under varying exercise levels. ECG signals are acquired via a wearable module and transmitted wirelessly to a processing device, then  Singular Value Decomposition (SVD) is applied to denoise the signals.  Extracted features are stored both locally and in the hospital database. During authentication, new ECG and motion data are collected, motion is detected, and the ECG is adaptively denoised then it is compared with stored ECG signal. An accuracy rate of about 90\% was achieved when the model was tested on the MIT-BIH Arrhythmia (MA) database and the MIT-BIH Noise Stress Test (NST) database.
 
In \cite{30}, a 1D-CNN model is proposed for identifying and classifying multiple simultaneous contaminants in sEMG signals without prior feature extraction. The model was evaluated using data collected from five subjects under five different contamination scenarios: ECG+MA+PL\footnote{Power Line noise}, ECG+AS+AWGN, ECG+MA\footnote{Motion Artifacts}+AWGN, ECG+PL+AWGN, and PL+AS+AWGN\footnote{ Additive White Gaussian Noise}. In all cases, the model achieved an accuracy exceeding 85\%.

The multi-layer Convolutional Neural Network (CNN) was proposed in \cite{31} for wireless ECG signals and motion sensors based biometric identification. Time and frequency domain features were extracted and used to train  the proposed model. The model was tested using data collected from a total of 34 participants who performed seven distinct activities while their physiological and motion data were recorded. The model obtain over 98\% classification accuracy, with 85\% of participants being identified with 100\% accuracy. Additionally, the model recognized the performed activities with an average accuracy of 92\%.


The proposed approaches demonstrate reliable detection of complex signal manipulations. However, most of these approaches account for external manipulations. In contrast, the present work does not rule out internal manipulation, i.e., the deliberate manipulation of ECG data by users to achieve various goals, such as concealing heart conditions and making insurance claims. In contrast to the data used in previous case studies, which were collected in controlled environments, the present work used data collected outside of clinical settings during various activities by subjects. Such data are difficult to evaluate and contain significant motion artifacts. We propose robust machine learning models to ensure that these challenges are successfully overcome.

\section{Data Acquisition}
\label{sec:data}
We employed the Shimmer platform (version 3)\footnote{\url{https://shimmersensing.com/product/consensys-ecg-development-kits/}.} to measure cardiac and physical activities. The platform avails 5 ECG channels, all of which were sampled synchronously at a rate of 512 samples per second. The measurements were taken in four separate batches of experiments. The first batch took place in 2019, with 8 healthy subjects (all males, mean age = 30 yrs, SD = 6 yrs) performing 7 different activities (sitting, standing, bending over, climbing up and down a staircase, jumping on the spot, walking, and running). Each activity lasted 120~s. The second and the third batches took place in 2024. The second batch consisted of 16 subjects, 11 of which were females and 5, males. For this batch, the mean age = 27 yrs and SD = 13 yrs. Thirteen of the subjects were healthy; one of them had asthma, another took regular medication which could affect blood pressure; and one of them, a 27 years old female, was a smoker. The third batch consisted of 10 healthy subjects, five females and 5 males, all between 21 and 24 years of age. The mean age was 22 yrs and SD = 1.9 yrs. The fourth batch took place in 2025 and consisted of 20 subjects. The average age in this batch is 24.2 with a SD = 2.26 yrs.

All data were collected with the permission of the TU Dresden's Ethic Committee  (under Application No. EK271072017). Full consent from all participants had  been obtained prior to the experiments. 

\begin{table}[H]
\centering
\caption{Summary of All Batches}
\label{tab:related_work_comparison2}
\begin{tabular}
{p{1.5cm}p{1.5cm}p{1.8cm}p{2cm}}\hline
\textbf{Batch (Year)} & \textbf{Number of Participants} & \textbf{Male/Female Ratio (M/ F)} & \textbf{Mean Age $\pm$ SD} \\
\hline
2019 & 8 & 8/0 & \(30 \pm 6\) years \\
2024 & 16 & 5/11 & \(27 \pm 13\) years \\
2024 & 10 & 5/5 & \(22 \pm 1.9\) years \\
2025 & 20 & 17/3 & \(24.2 \pm 2.26\) years \\\hline
\end{tabular}
\end{table}
\section{Features}
\label{sec:features}
Preprocessing of the ECG signal is conducted to extract features  effectively by removing artifacts, reducing noise, and standardizing inputs. The pre-processing pipeline includes signal segmentation, band-pass filtering, normalization, and transformation input into the time-frequency domain.
\subsection{Segmentation}
The raw ECG signals is segmented into fixed-length frames of 4 seconds (corresponding to 2048 samples at a 512 Hz sampling rate), with a 30\% overlap between consecutive segments. Segments shorter than 4 seconds were excluded from further processing. The 30\% overlap is chosen to enhance the detection of tampering patterns that occur near segment boundaries  while avoiding excessive data redundancy and computational overhead. To have sufficient  information from ECG  signal for robust wave form analysis and maintaining a manageable input size,  a 4-second segment length was selected. Segments shorter than  this risk omitting critical signal features, whereas segment
longer than  these windows may increase computational demands and dilute the localized detection of tampering artifacts.
\subsection{ Filtering}
The raw ECG signals are contaminated with various types of noise, including baseline wander due to respiration, power line interference at 50\slash 60 Hz, and high-frequency noise originating from muscle activity.  A  band-pass second-order Butterworth filter with cutoff frequencies set at 0.5 Hz and 100 Hz is used to effectively attenuate noise while preserving clinically relevant cardiac signal feature.  This frequency was selected since frequencies below 0.5 Hz typically correspond to baseline drift caused by respiration and electrode motion, whereas frequencies above 100 Hz are  frequencies of muscle artifacts and external interference.
\subsection{Min-Max Normalization}
After conducting  the filtering process , Min-Max normalization is applied to each ECG segment, except the segments used for the identity verification task, the segment in the identity verification task is not normalized in order to preserve the original morphological differences between subjects.

To ensure that tampered regions appeared subtle and visually indistinct, particularly at the transition points between authentic and manipulated segments, each segment was independently normalized to a [0, 1] range.  In scenarios involving partial tampering, signal segments from different sources may exhibit varying baseline offsets or amplitude scales and can create detectable transitions if the ECG signal is not normalized properly.

\begin{figure}[H]
    \centering
    \begin{subfigure}[b]{0.48\textwidth}
        \centering
    \includegraphics[width=\textwidth]{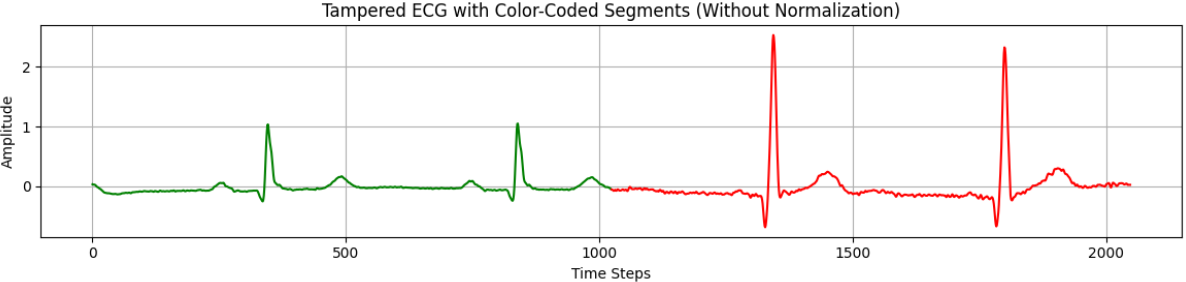}
        \caption{before normalization}
        \label{fig:figf1}
    \end{subfigure}
    \hfill
    \begin{subfigure}[b]{0.48\textwidth}
        \centering
        \includegraphics[width=\textwidth]{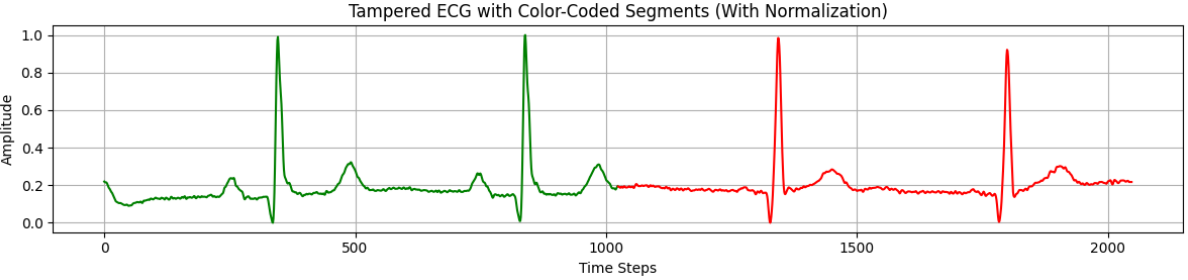}
        \caption{ after normalization}
        \label{fig:figf2}
    \end{subfigure}
    \caption{Tampering before and after normalization}
\label{fig:ecg_comparison}
\end{figure} 
\subsection{Continuous Wavelet Transform (CWT)
}The scalograms that  captured both temporal and spectral characteristics of the ECG signals are generated by using the Continuous Wavelet Transform (CWT) method,  a method used to transform the one-dimensional ECG time-series signals into two-dimensional time-frequency signals. Each segment produced a transformed output of shape (2048, 96), where 2048 corresponds to time samples and 96 to frequency bins. The choice of 96 scales provided a balance between resolution and computational efficiency.
These generated scalograms served as inputs for the hybrid model (transformer-CNN), and the original preprocessed 1D ECG signals were used as an input for  the CNN and ResNet models. The hybrid model used the generated input signal, since it is difficult to   extract localized frequency variations from  the raw time-domain ECG signal. 
\subsection{Tampering ECG Signal}
There are two types of tampering, the first one is partial segment tampering, a tampering mechanism where  some segments span the boundary between original and injected data. The second one is a full segment replacement tampering mechanism, where the entire segments  fall within tampered regions and appear structurally coherent, resembling genuine ECG signs.
These two tampering mechanisms have  distinct characteristics, and therefore  it is difficult to detect both tampering scenarios effectively by using a single method.
To address this challenge, two detection strategies are proposed:
Transformer, CNN, ResNet and Transformer-CNN hybrid models are trained to identify structural anomalies linked to partial tampering.
For identity verification (full segment replacement), a Siamese neural network was used to verify the identity associated with each segment. 
In this paper, to evaluate the performance of the model proposed, different types of  tampering ECG signals are generated artificially, and also various partial tampering techniques were designed with differing levels of complexity and subtlety. All  manipulation methods are applied at the segment level, with each ECG segment spanning 4 seconds (2048 samples at a 512 Hz sampling rate).
Each tampering scenario was formed by merging an ECG segment from one person (person A) with other individual (person B). The  ECG segments are recorded from each person when they are doing the same physical activity;  this activity-aware pairing strategy is adopted to avoid bias of the  model toward detecting changes in activity rather than actual signal level inconsistencies indicative of tampering. Six tampering scenarios were simulated by using  the blending method.

 \begin{figure}[H]
    \centering
    \includegraphics[width=0.48\textwidth]{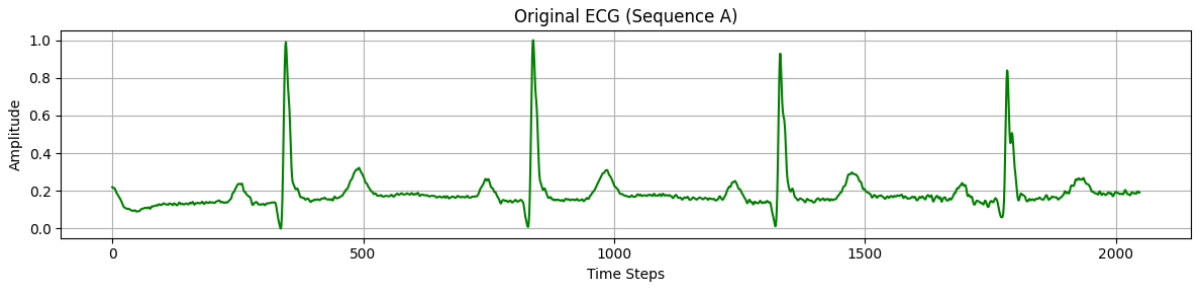}
    \caption{Original ECG from Person A}
\end{figure}

\begin{figure}[H]
    \centering
    \includegraphics[width=0.48\textwidth]{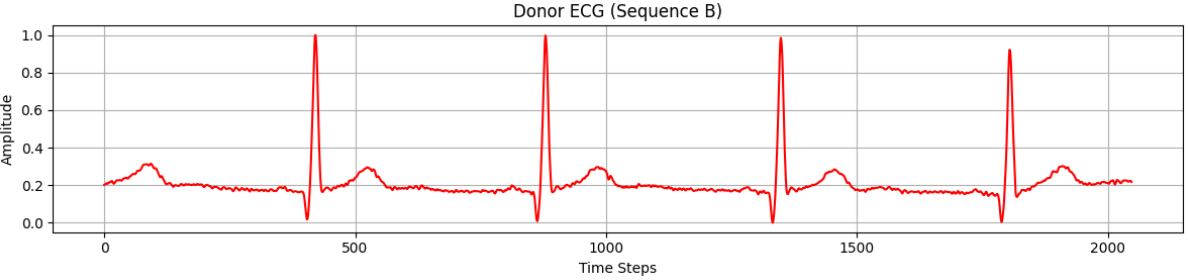}
    \caption{Donor ECG from Person B}
\end{figure}
Blending Process:
Linear blending method is proposed to achieve a smooth transition between the end of the preceding segment and the beginning of the subsequent one,  it is used to ensure a gradual transition by weighting signal values across a fixed blending window (e.g., 5 samples), thereby reducing the visibility of abrupt amplitude or phase changes that could reveal tampering artifacts.
\[
\text{blended} = (1 - \alpha) \cdot \text{prev\_tail} + \alpha \cdot \text{curr\_head}\] Where $\alpha$ is a linearly spaced vector from 0 to 1 across the blending width.
\begin{figure}[H]
    \centering
\includegraphics[width=0.48\textwidth]{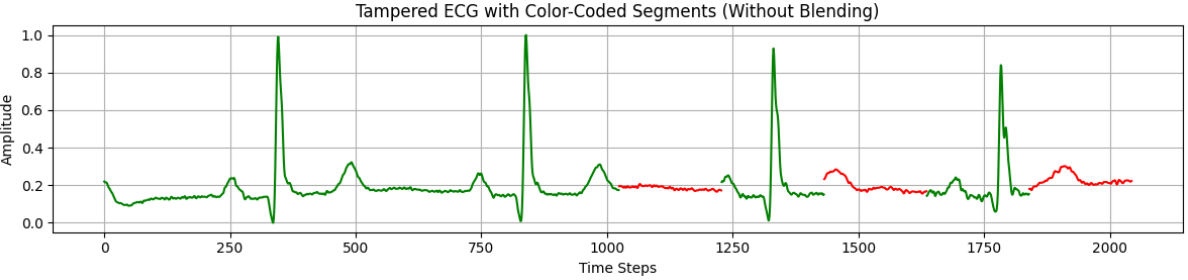}
\caption{Tampered ECG segment without linear blending.}
\label{fig:tamper_without_blending}
\end{figure}

\subsubsection{50-50 Replacement}
In this tampering scenario, 50\% of the ECG segment is from person A, and the remaining 50\% is from person B. This tampering  mechanism is one of the simplest tampering mechanisms,  but the use of linear blending for merging the two ECG signals  at the midpoint makes it difficult to be detected.

\begin{figure}[H]
    \centering
\includegraphics[width=0.48\textwidth]{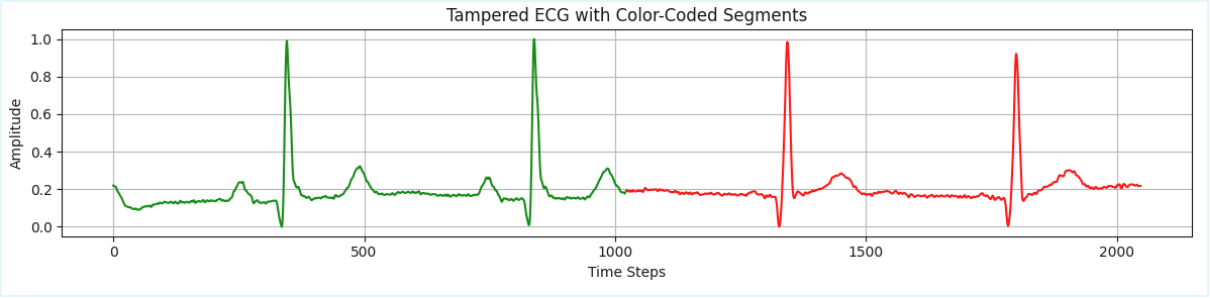}
    \caption{Tampered ECG: First half from Person A (green), second half from Person B (red)}
\end{figure}
\subsubsection{75-25 Asymmetric Tampering (A-B)}
This tampering strategy keeps 75\% of the segment from Person A and substitutes the last 25\% with Person B’s signal. The goal is to produce a more subtle tampering that influences a trailing part of the segment. The segment boundary is close to the end, and because of blending, it is hard to see and to detect algorithmically.
\begin{figure}[H]
    \centering
\includegraphics[width=0.48\textwidth]{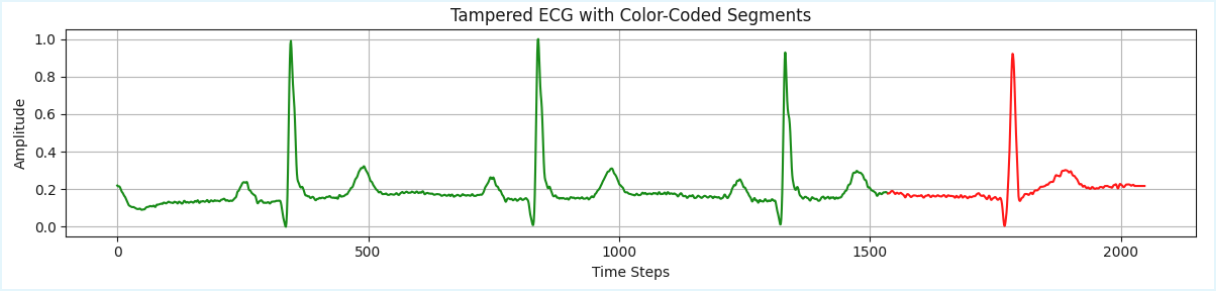}
\caption{Tampered ECG: 75\% from A, 25\% from B}
\end{figure}
\subsubsection{50-25-25 Multi-Source Replacement (A-B-A)}
This tampering mechanism used a three part composition, where the first 50\% of the signal is from Person A,   25\% from Person B and the remaining  25\% from Person A. 
\begin{figure}[H]
\centering
\includegraphics[width=0.48\textwidth]{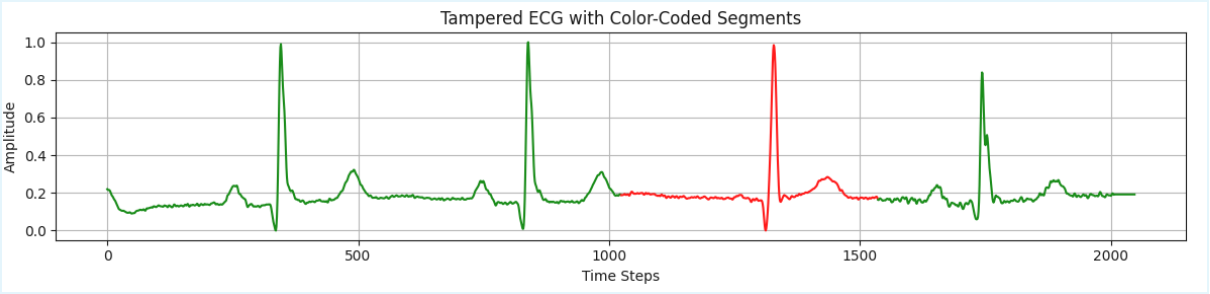}
\caption{Tampered ECG: A-B-A pattern, with colour-coded segments}
\end{figure}
\subsubsection{Alternating Blend Pattern (50-10-10-10-10-10)} This tampering method was highly structured, where the attacker attempts to make the data look more natural by constantly alternating the ECG signals From person A  and person B. The combination mechanism used in this tampering was the  first half was  from Person A,  10\% from B, and then 10\% from A, and for the remaining 30\%  it switches between A and B.  This type of tampering is rhythmic tampering, which is challenging to detect visually because of quick transitions and uniform blending at every transition. 
\begin{figure}[H]
\centering
\includegraphics[width=0.48\textwidth]{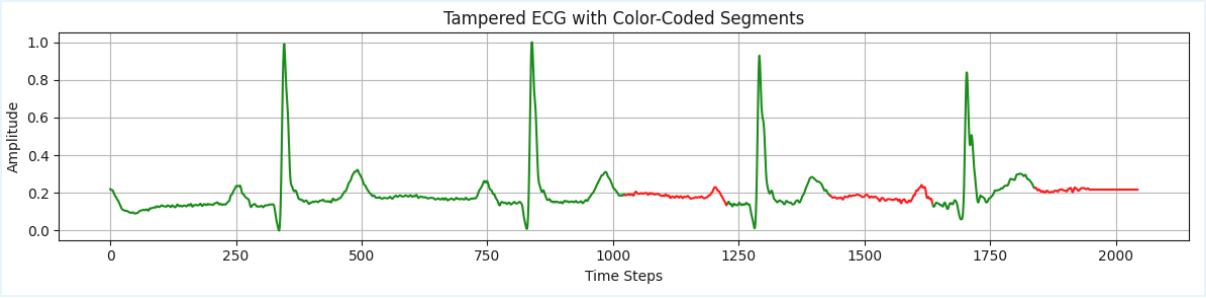}
\caption{Tampered ECG: Alternating blend of Person A and B (color-coded)}
\end{figure}
\subsubsection{Sporadic 20\% Replacement} In this tampering method, 20\% of the total ECG segment is substituted by several short, non-contiguous donor fragments of equal size. In particular, four segments, each accounting for 5\% of the total length of the sequence, are placed at random locations in the 4-second signal.  No large contiguous area is entirely from Person B, making the tampering subtle and hard to notice visually or statistically. 
\begin{figure}[H]
\centering
\includegraphics[width=0.48\textwidth]{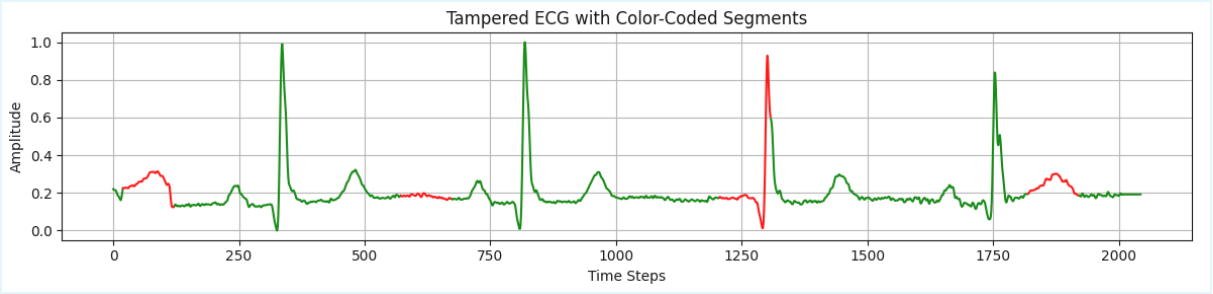}
\caption{Tampered ECG: Sporadic insertions from Person B (red) into Person A’s signal (green)}
\end{figure}
\subsubsection{Sporadic 50\% Replacement} This strategy expands the 20\% variant by raising the tampered part to 50\% of the
signal. The segment is interrupted by ten donor fragments, each of which make up 5\%
of the ECG.  The fact that this tampering is common,
but granular, might present a unique challenge to detection models, because the
changes are subtle, but pervasive across a large amount of the data.
\begin{figure}[H]
\centering
\includegraphics[width=0.48\textwidth]{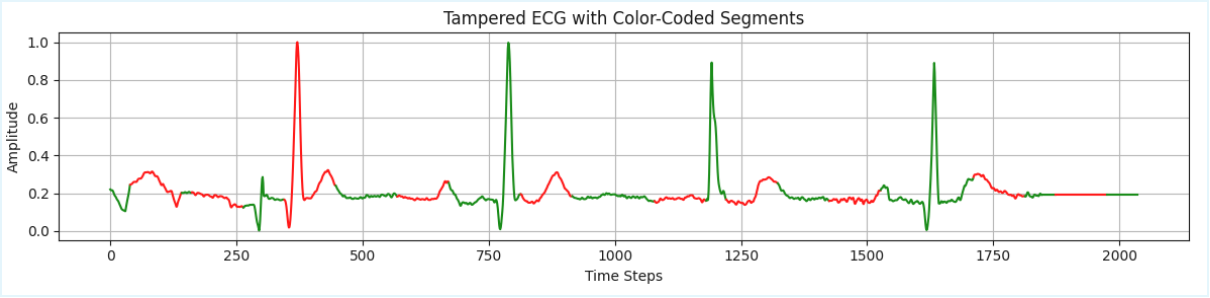}
\caption{Tampered ECG: Roughly half of the segment replaced with donor signal (red)}
\end{figure}

\section{Model}
\label{sec:model}
This section outlines  the deep learning models we developed for tampering detection and person identification.  The models developed for tampering detection identify  signal irregularities that indicate manipulation, and the models for individual identification determine whether a given segment belongs to a particular individual.

\subsection{Model Architectures for Partial Tampering Detection}
\subsubsection{CNN Model}
The CNN model developed  detects tampering artifacts from one-dimensional ECG signals by extracting  short and localized temporal patterns like QRS complexes, P-wave disruptions and small frequency abnormalities, which are important tampering indicators in ECG signals.  

The model proposed is constructed from 3 Conv1D layers followed by batch normalization, dropout and max pooling. To enable the model to first capture a larger temporal pattern and then proceed with capturing  finer local patterns,  the convolutional layers are formed with increasing filter sizes of 64, 128, and 256, and decreasing kernel sizes of 7,5 and 3, respectively. All convolutional layers have ReLU activation and same padding so that temporal dimensions are preserved before pooling. One-dimensional MaxPooling layers with a pool size of 2 are used to down-sample the temporal resolution, and dropout layers with a rate of 0.3 are used after each convolutional block for regularizing training. After the last convolution layer, the feature maps are flattened and forwarded into a fully connected classification head that has two dense layers with 128 and 64 nodes and  ReLU activation. The last output node employs a sigmoid activation for a binary ECG sequence classification of clean and tampered ECG sequences.

\begin{figure}[h]
  \centering
\includegraphics[width=0.45\textwidth]{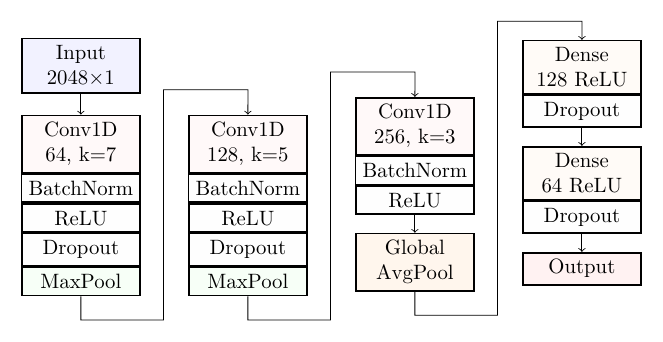}%
  \caption{1D Convolutional Neural Network Model}
  \label{fig:cnnpd}
\end{figure}

\subsubsection{ResNet Model}

The residual network (ResNet) architecture adopted in this study builds on the basic CNN design by adding residual connections, so that the model can learn deeper feature representations without suffering from the problem of gradient vanishing. The ResNet is made up of several stacked residual blocks that have two Conv1D layers with Batch normalization and ReLU activation functions.  The residual connection skips over these two layers and is simply added to the block’s output. After the residual blocks, the model uses a global average pooling layer to compact the temporal dimensions giving a fixed size feature vector regardless of the length of the input sequence. Such representation is then fed to a dense layer with a sigmoid activation function that does binary classification by classifying tampered and clean
ECG segments. Stacking several residual blocks gives the ResNet the depth to extract hierarchical temporal features at different resolutions. This allows the model to identify tampering artifacts that can appear in various temporal scales from small waveform distortions to longer disruptions. With the help of residual learning, the architecture preserves high representational power with computational efficiency, which makes the architecture suitable for the tampering detection task on one-dimensional ECG inputs.

\begin{figure}[h]
\centering
\begin{subfigure}[b]{0.45\textwidth}
\includegraphics[width=\textwidth]{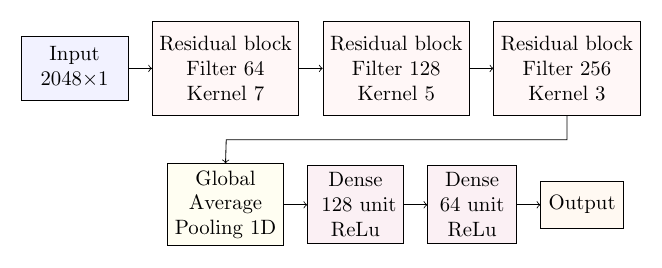}
\caption{1D ResNet Model}
\label{fig:Resnet}
\end{subfigure}
\hfill
\begin{subfigure}[b]{0.45\textwidth}
\includegraphics[width=\textwidth]{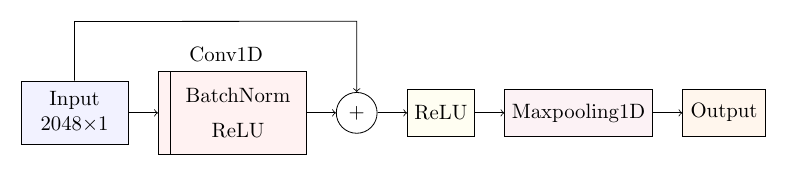}
\caption{Internal structure of the residual block used in the ResNet.}\label{fig:resnet1}
\end{subfigure}
\caption{Detailed view of the ResNet model architecture and its residual block design.}
\label{fig:resnet block}
\end{figure}

\subsubsection{Transformer Model} Transformer-based architectures are a key focus of this work. Transformer-based models can learn temporal dependencies in long ECG segments and can therefore be highly effective at detecting ECG tampering compared to CNN models, where inconsistencies can be subtle and spread across multiple cardiac cycles. This section presents various transformer-based models used in this study.
\\  \\
\textbf{i. Pure Transformer with Deep Feed-Forward Layers:}

This  model operates with  ECG segments of shape 2048$\times$96, which correspond to 4 s of ECG data in the time–frequency domain.  Positional encoding is applied to the input to enable the model to learn key temporal placements of the waveforms (P-wave, QRS complex, and T-wave), where morphology and timing are important for anomaly detection.

In this model, three stacked Transformer encoder blocks are selected to balance model complexity and generalization.  Each encoder block has a multi-head attention block consisting of 8 attention heads, each of which has a subspace of 48 dimensions. Such a configuration allows the model to concentrate on several aspects of the ECG signal at once, including short-term variations and long-term dependencies of the waveform. 

After the self-attention layer in every encoder block, a deep feed-forward network (FFN) is used to increase the representational ability of the model. Unlike the conventional two layer design, this FFN has three dense layers with a gradually reduced dimensionality, 384 units (4$\times$96), 192 units (2$\times$96), and 96 units. The first expansion to 384 units allows richer feature transformations, extracting complex interactions from the attention outputs. The stepwise decrease to 96 units filters and fine-tunes these features while preserving compatibility with residual connections, allowing stable training dynamics.
GELU activations post the first two layers implement smoothing for fine-grained ECG fluctuations, and dropout layers after each dense operation reduce overfitting and enhance generalization to unobserved tampering patterns. A global average pooling operation is performed along the temporal dimension after the three transformer encoder blocks. This aggregation step creates a compact feature vector that summarizes the information over all 2048 time steps. Then two more dense layers of 512 and 256 units are used to further fine-tune the feature representation to produce the final classification output.

\begin{figure}[H]
\centering
\includegraphics[width=0.45\textwidth]{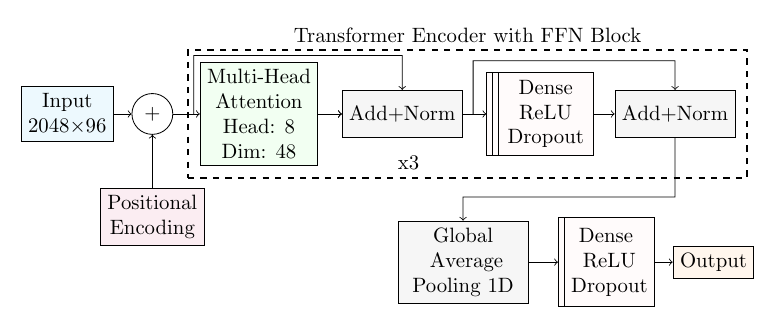}
\caption{Transformer Block with deep Feed-Forward Network and FFN classifier}
\label{fig:TrFFN2}
\end{figure}

\textbf{ii. Transformer with CNN as Feed-Forward Layers:}

This model is developed by replacing  the feed-forward network (FFN) block in Fig.~\ref{fig:TrFFN2} with a CNN-based block. Except for this modification, the rest of the architecture positional encoding, multi-head attention, and global average pooling is consistent with the Transformer Block described above. The convolutional layers are constructed with successively reducing kernel sizes of 7, 5, and 3 and successively increasing  filter sizes of 64, 128, and 96. By decreasing the kernel size along layers, the model constructs a hierarchy from coarse- to fine-grained temporal patterns. The more filters available on the convolutional layers, the more expressive the feature extraction can become with the increasing localization of the receptive field. Batch normalization is used after every convolutional operation to ensure that learning is stable. Dropout layers are also added after each convolution in order to avoid over fitting. This approach combines the strengths of local pattern extraction of CNNs and the long-range dependency learning capability of transformer models. 

\begin{figure}[H]
\centering
\includegraphics[width=0.45\textwidth]{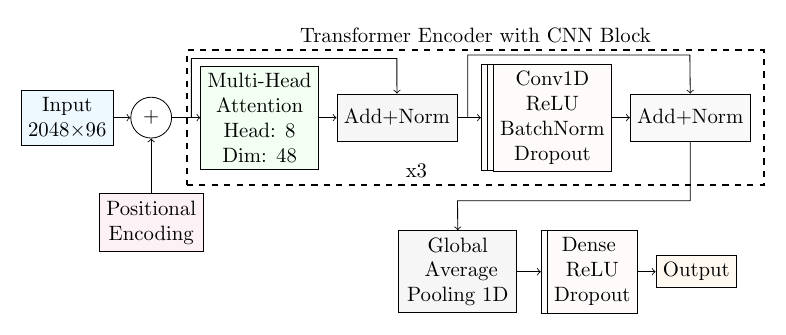}
\caption{Transformer block with CNN feed-forward layers and FFN classifier}
\label{fig:TrCNN3}

\end{figure}\textbf{iii. CNN as Feature Extractor $+$ Transformer with deep FNN:}

To achieve better tampering detection from  raw one-dimensional ECG segments of shape 2048$\times$1, a hybrid structure of convolutional feature extractor and Transformer encoder is proposed. 
The CNN model used to extract features is set up with step-down Kernel sizes at values of 13, 5, and 3 and  with corresponding step-up filter depths of 64, 128 and 256 respectively. The use of a bigger kernel in the first layer enables the model to learn wide temporal trends that are then refined into localized morphological features by  smaller kernels in subsequent layers. one-dimensional MaxPooling operations with pool size 2 are applied to decrease the temporal resolution and emphasize the dominant patterns, while Batch Normalization and Dropout are applied all along to provide training stability and regularization. After feature extraction, the output is fed into sinusoidal positional encoding to restore the awareness of temporal order by the model.  The encoded features are then passed through 3 Transformer encoder blocks, which are the same as the Transformer encoder blocks described in Fig.~\ref{fig:TrFFN2}.  This design enables the model to first convert the raw ECG signals into local feature representations and then globally reason over these features using the Transformer model. The results showed that it is possible to operate on one-dimensional ECG inputs directly and apply deep sequential modeling to achieve a significant enhancement of the detection of tampering artifacts.\\

\textbf{iv. CNN Feature Extractor$+$Transformer with CNN Feed-Forward:}

In this version, the input is raw 1D ECG segments of shape 2048$\times$1. The CNN  model used to extract features  is similar to the CNN feature extractor used above, except the first kernel size is changed from 13 to 7. After the feature extraction and positional encoding, the features are fed to a Transformer encoder, which is the same as the Transformer encoder used in Fig.~\ref{fig:TrCNN3}. The idea behind this architecture is to add hierarchical local feature refinement at various stages of the network: first at the raw signal level through the CNN feature extractor and then at the intermediate feature level through the CNN-augmented feed-forward stages of the Transformer. By the use of convolutional operations both before and within the Transformer blocks, the model is set to detect faint tampering signatures, which can either appear as localized irregularities in the waveform or as wider sequential discrepancies in the ECG signal. \\

\textbf{v. CWT-based CNN Feature Extractor $+$Transformer:}

This model is developed by extending the architecture shown in Fig.~\ref{fig:TrFFN2}, a CNN module for feature extraction is incorporated and  ECG signals with size 2048$\times$96 is used as an input. The CNN feature extractor is constructed as three sequential Conv1D layers, where the filter depth increases (64, 128, and 256) and the kernel size decreases (7, 5, and 3). In this model, batch normalization, ReLU activations, and dropout layers are used to minimize overfitting, and one-dimensional MaxPooling layers are used  to improve the extraction of key features. Then positional encoding is added to the feature sequence  to model the sequential nature of the cardiac waveforms. The sequence is then sent through a set of Transformer encoder blocks. Unlike other architectures relying on deep feed-forward networks, this model uses a simple two-layer FFN in its Transformer block. This FFN first expands the feature dimension to 1024 units using a GELU activation and then projects it back to 256 units. Once the Transformer has processed the sequence of feature vectors, global average pooling is used to create a fixed-size feature vector. The feature vector is processed by two dense layers, each with 512 and 256 units, before being used to generate a binary output for detecting tampering. In comparison to other models, this model has two major differences: Firstly, the model uses CWT-transformed 2D ECG inputs rather than raw 1D signals; secondly, it uses a regular Transformer encoder rather than deep or CNN-based FFNs. By using CNN feature extraction on the CWT spectrograms, the model can learn discriminative spatial patterns that are characteristic of anomalies. The Transformer encoder is responsible for modeling temporal relations between such features extracted so that detection of fine or distributed patterns of tampering that appear over time can be done.

\subsubsection{Training and Evaluation for Tampering Detection Models}

All models were trained by using a balanced mix of tampered and untampered ECG signals, for each experiment, 50\% of the sequences were clean, while the remaining 50\% were tampered.  The data were segmented into 4-second windows with 30\% overlap to increase dataset density. To avoid artifacts from activity misalignment, tampering was performed only between segments representing the same physical activity (e.g., walking segments from one participant were replaced with walking segments from another). The experimental inputs consisted of either continuous wavelet transform  for Transformer-based variants or raw 1D ECG signals for models where CNNs served as the initial feature extractor.



\subsection{Computational Efficiency of Models}
Floating Point Operations (FLOPs) measure the overall number of arithmetic operations (multiplications and additions) that are required by a model in a single forward pass. In this paper, as shown in table ~\ref{tab:computational-effort}, FLOPs are used to compare the computational burden of various tampering detection and person verification architectures quantitatively, so that the evaluation is not skewed towards accuracy but also the resource-efficiency and the implementation and deployment complexities of the models.

\begin{table}[H]
\centering
\caption{Model Computational Effort and Input Size in million FLOPS} 
\label{tab:computational-effort}
\begin{tabular}{lll}
\toprule 
\textbf{Model} & \textbf{Input Size} & \textbf{FLOPs} \\ \midrule 
CNN & 2048$\times$1 & 288 \\
ResNet & 2048$\times$1 & 728 \\ 
Tran-DeepFFN & 2048$\times$96 (CWT) & 4646 \\
Tran-CNNFFN & 2048$\times$96 (CWT) & 3926 \\
FeatCNN-Tran & 2048$\times$1 & 4277 \\
FeatCNN-TranCNN & 2048$\times$1 & 4244 \\
CWT-FeatCNN-Tran & 2048$\times$96 (CWT) & 2179 \\ 
\bottomrule 
\end{tabular} 

\end{table}

\begin{table*}
\centering
\caption{Detection Accuracy (\%) for Different Tampering Strategies}
\label{tab:DetectionAccuracy}
\begin{tabular}{ p{4cm}p{1.5cm}p{1.5cm}p{1.5cm}p{1.5cm}p{1.5cm}p{1.5cm} }
\hline
\textbf{Model} & \textbf{50-50} & \textbf{75-25} & \textbf{50-25(2)} & \textbf{50-10} & \textbf{Sp20} & \textbf{Sp50} \\
\hline
Tran-Deep FFN & 85.50& 85.62 & 87.14& 85.22& 91.33& 98.41\\
Tran-CNN FFN & 86.50& 85.80& 87.30& 92.10 & 91.62& 98.91 \\
FeatCNN-Tran & 98.07 & 98.54 & 99.81& 100& 99.89& 99.92 \\
FeatCNN-TranCNN & 98.46& 98.68 & 99.78 & 100& 99.83& 99.94\\
CWT-FeatCNN-Tran & 94.90& 95.48 & 95.50& 98.54 & 99.36 & 99.81\\
CNN & 97.90 & 98.30 & 99.47& 99.59& 99.80& 99.90 \\
ResNet & 98.10& 98.78& 98.36 & 100& 99.90 & 100\\
\hline
\end{tabular}
\vspace{0.3cm}
\footnotesize\\
{Note: Sp20 = Sporadic 20\%, Sp50 = Sporadic 50\%, 50-10 = 50\% A + 10\% B alternating, 50-25(2) = 50\% A + 25\% B + 25\% A, 50-50 = 50\% A + 50\% B, 75-25 = 75\% A + 25\% B.}
\end{table*} 

\section{Evaluation}
\label{sec:evaluation}
For tampering identification, seven models are proposed, and  the performance of each model is evaluated with the six tampering scenarios. Stratified sampling is used to split the data collected into 80\% training, 10\% validation, and 10\% testing sets. Each experiment is repeated 25 times, and the reported results represent the average accuracy over these runs.

\subsection{Tampering Detection Results}

As shown in Table,~\ref{tab:DetectionAccuracy}, all models performed well in the classification of heavily tampered data, especially in the scenarios with sporadic 20\% or 50\% tampering, where detection accuracies varied from 91\% to 98\% respectively. The performance differences between the individual models became more apparent in more subtle cases such as 50:50 and 75:25 segment swaps. The models combining convolutional feature extraction with transformer encoding achieved some of the best results. Interestingly, ResNet was consistently competitive in all scenarios, though the transformer-based models exhibited the smallest performance variance across the 25 independent experiments. The CNN-based models exhibited significant performance fluctuations.

To further gain insights about the  performance of each model, the performance matrices  were separately analyzed for each model and the average values are summarized  in Tables ~\ref{tab:50alt-metrics}. Accordingly, the FeatCNN-TransCNN and FeatCNN-Trans models obtained the highest accuracies of 99.45\% and 99.37\%, respectively.

\begin{table}
\centering
\caption{The average values of  accuracy, precision, recall, and
F1-score for all Tampering scenario}
\label{tab:50alt-metrics}
\begin{tabular}{
p{2.5cm}p{0.9cm}p{0.9cm}p{0.8cm}p{0.7cm}}
\toprule
\textbf{Model} & \textbf{Accuracy } & \textbf{Precision } & \textbf{Recall } & \textbf{F1-Score } \\
\midrule
Tran-DeepFFN & 88.87 & 87.84 & 89.88 & 88.85 \\
Tran-CNNFFN & 90.37 & 89.5& 91.23 & 90.35\\
FeatCNN-Tran & 99.37 & 99.33 & 99.4& 99.37\\
FeatCNN-TranCNN & 99.45 & 99.49 & 99.41& 99.45 \\
CWT-FeatCNN-Tran & 97.26& 96.91 & 97.57 & 97.24 \\
CNN & 99.16 & 98.92 & 99.37 & 99.14 \\
ResNet & 99.19 & 98.82& 99.66 & 99.16 \\\bottomrule
\end{tabular}
\end{table}



\subsection{Training Stability Observations}

In a highly localized tampering scenario, the CNN and ResNet models achieved  high performance, nevertheless, they were sensitive to training parameter variations, exhibiting inconsistent training results. Minor variations in training configurations often led to significant performance differences. Therefore, they required careful hyperparameter tuning and stabilization to achieve consistent results.

\begin{table*}[hbt]
\centering
\scriptsize
\caption{Compact comparison of ECG based biometric verification and tampering detection related studies}
\label{tab:related_work_comparison1}
\begin{tabular}{|p{0.5cm}|p{2.8cm}|p{1.2cm}|p{1.3cm}| p{1.5cm}|p{2.3cm}|p{1.3cm}|p{2.5cm}|}
\hline
\textbf{Work} & \textbf{Dataset} &\textbf{Environment}&\textbf{No. of subject}\textbf&\textbf{Model}& \textbf{Performance}&\textbf{Identification\slash tampering} & \textbf{Remarks}   \\
\hline
\cite{24} & MIT-BIH, European ST-T,  Holter  & Clinical data set &47,N/A, N/A &HQCNN,  CNN& Accuracy(\%) HQCNN:94.3  CNN:92.5& yes\slash yes&  Relies on quantum computing    \\
\hline
\cite{25} & PTBDB, CEBSDB & Public dataset &290,  20&  MHT&EET: 7.62\% &yes\slash yes & The MHT can be manipulated by changing user-specific keys\\
\hline
\cite{26} &  FECGSYN,  ABFECGDB & Clinical dataset&10, 5&DAW-Net & F1 score of 98.13\% &no\slash yes&  Need accurate QRS localization\\
\hline
\cite{27} & MIT-BIH,  PTB & Public dataset&47, 290& CNN & Accuracy(\%) PTB: 99.15; MIT-BIH :98.5 &yes\slash no & Effective for  stable heart rhythms\\
\hline
\cite{28} &  Fantasia, MITDB, ECG-ID, EDB, ADE  &Public dataset &40, 47, 90, 79, 25& SAFD & Accuracy: 97.59\%&yes\slash no& Sensitive to noise \\
\hline
\cite{29} & MIT-BIH Arrhythmia,  self-generated ECG dataset&Public dataset, uncontrolled &47,30& SVD & Accuracy: 90\% &yes\slash no & Existence of Linear correlation assumption  \\
\hline
\cite{30} & Self-generated ECG dataset  & Uncontrolled&5& 1D-CNN & Accuracy: 85\%&yes\slash yes& Limited data set\\
\hline
\cite{31} & Self-generated ECG dataset &Uncontrolled&34& CNN& Accuracy: $>$92\% & yes\slash no & Data collected with 7 different activities \\
\hline
\rowcolor{LightGray}
{This work} & Self-generated ECG dataset &Uncontrolled&54&ResNet, CNN, transformer+CNN  & Accuracy:  $>$ 98\% &yes\slash yes&Multiple tampering scenarios are considered\\
\hline
\end{tabular}
\end{table*}

\subsection{Discussion}

The experiment conducted demonstrate that FeatCNN-Tran, FeatCNN-TranCNN and ResNet models could  effectively detect highly fragmented distortion introduced by tampering, even though such distortions could be easily mistaken for physiological or environmental noise.  In particular, the intelligent tampering methods (50\%-50\% and 75\%-25\% replacement) introduced subtle changes in time-domain and frequency-domain features, however these modifications were successfully identified by FeatCNN-TranCNN model with notable accuracy exceeding 98\%. This said, the transformer-based models incurred a considerable computation cost, as can be seen in Table~\ref{tab:computational-effort}.

Compared to prior studies that focus on detection of external adversarial manipulation and ECG based biometric identification, this study addresses less-explored yet critical challenge: the Intentional self-alteration and sophisticated modification of ECG signal for malicious purposes. In addition, the present work provides more advanced and robust tampering detection models by taking into account the strengths of different ML models. The study proposed improve the integrity and security of remote health monitoring systems. Table ~\ref{tab:related_work_comparison1} summarizes the body of work reviewed in section~\ref{sec:related} and the study conducted in this article. The table highlights the importance of protecting the integrity of biomedical measurements. While many of the proposed approaches use publicly available datasets generated in controlled environments to develop and test their models, some rely on self-generated datasets created outside of clinical settings. Interestingly, the performance of these approaches is not inferior to those based on datasets created in controlled environments. Our approaches fall into the second category. However, our tamper emulation and detection models are significantly more robust. Furthermore, the data acquisition process involves more complex activities, which significantly impact the quality of the ECG measurements provided to the models. Nevertheless, the accuracy they achieved for tampering detection and person identification is better than the state of the art.

\section{Conclusions}
\label{sec:conclusions}
This research evaluated the performance of CNN, ResNet, and  transformer-based hybrid models for ECG tampering detection. Experimental results show that in highly fragmented manipulation scenarios, CNN, FeatCNN-TranCNN, FeatCNN-Tran and ResNet models achieved above 99.5\% accuracy; for subtle manipulations (e.g., 50\%-50\%) and (75\%-25\% replacement), the FeatCNN-TranCNN models performed consistently better, achieving above 98\% accuracy. The performance of  CNN-based models (standard CNN and ResNet) is high  in moderately tampered conditions; however, they exhibited high variability during training, indicating sensitivity to hyperparameters and a tendency toward overfitting.  In contrast, the hybrid model achieved high performance and also demonstrated more stable training across varying tampering levels. Pure Transformer models (Tran-DeepFFN) that work on CWT-transformed inputs performed relatively poorly in terms of accuracy, particularly in cases of subtle tampering scenarios. Additionally, CWT-FeatCNN-Tran; where the inputs are first modified by CWT, achieved moderate accuracies (about 95\%-96\%). It is also worth mentioning that CWT transformations also added additional computational overhead at the pre-processing step, and higher-dimensional CWT inputs (2048×96) increased model complexity and inference time. Hybrid designs that incorporated CNN components for feature extraction or into the Transformer feed-forward network performed better in terms of finding balance between stability, computational efficiency, and peak accuracy. Future work will focus on incorporating more dataset to include varied patient profiles and integrating the models with edge computing platforms to enable real-time detection in wearable and mobile devices.

\balance
\bibliographystyle{IEEEtran}
\bibliography{library}

\begin{thebibliography}{10}
\providecommand{\url}[1]{#1}
\csname url@samestyle\endcsname
\providecommand{\newblock}{\relax}
\providecommand{\bibinfo}[2]{#2}
\providecommand{\BIBentrySTDinterwordspacing}{\spaceskip=0pt\relax}
\providecommand{\BIBentryALTinterwordstretchfactor}{4}
\providecommand{\BIBentryALTinterwordspacing}{\spaceskip=\fontdimen2\font plus
\BIBentryALTinterwordstretchfactor\fontdimen3\font minus
  \fontdimen4\font\relax}
\providecommand{\BIBforeignlanguage}[2]{{%
\expandafter\ifx\csname l@#1\endcsname\relax
\typeout{** WARNING: IEEEtran.bst: No hyphenation pattern has been}%
\typeout{** loaded for the language `#1'. Using the pattern for}%
\typeout{** the default language instead.}%
\else
\language=\csname l@#1\endcsname
\fi
#2}}
\providecommand{\BIBdecl}{\relax}
\BIBdecl

\bibitem{Sumalatha2024Deep}
U.~Sumalatha, K.~K. Prakasha, S.~Prabhu, and V.~C. Nayak, ``Deep learning
  applications in ecg analysis and disease detection: An investigation study of
  recent advances,'' \emph{IEEE Access}, 2024.

\bibitem{Abubaker2023Detection}
M.~B. Abubaker and B.~Babayiğit, ``Detection of cardiovascular diseases in ecg
  images using machine learning and deep learning methods,'' \emph{IEEE
  Transactions on Artificial Intelligence}, vol.~4, no.~2, pp. 373--382, 2023.

\bibitem{Behfar2021Fully}
M.~H. Behfar, D.~D. Vito, A.~Korhonen, D.~Nguyen, B.~M. Amin, T.~Kurkela,
  M.~Tuomikoski, and M.~Mäntysalo, ``Fully integrated wireless elastic
  wearable systems for health monitoring applications,'' \emph{IEEE
  Transactions on Components, Packaging and Manufacturing Technology}, vol.~11,
  no.~6, pp. 1022--1027, 2021.

\bibitem{kassem2021context}
A.~Kassem, M.~Tamazin, and M.~H. Aly, ``A context-aware iot-based smart
  wearable health monitoring system,'' in \emph{2020 international conference
  on communications, signal processing, and their applications (ICCSPA)}.\hskip
  1em plus 0.5em minus 0.4em\relax IEEE, 2021, pp. 1--6.

\bibitem{Uwaechia2021Comprehensive}
A.~N. Uwaechia and D.~A. Ramli, ``A comprehensive survey on ecg signals as new
  biometric modality for human authentication: Recent advances and future
  challenges,'' \emph{IEEE Access}, vol.~9, pp. 97\,760--97\,802, 2021.

\bibitem{Ribeiro2018Evolution}
J.~Ribeiro~Pinto, J.~S. Cardoso, and A.~Lourenço, ``Evolution, current
  challenges, and future possibilities in ecg biometrics,'' \emph{IEEE Access},
  vol.~6, pp. 34\,746--34\,776, 2018.

\bibitem{sassi2017pdf}
R.~Sassi, R.~R. Bond, A.~Cairns, D.~D. Finlay, D.~Guldenring, G.~Libretti,
  L.~Isola, M.~Vaglio, R.~Poeta, M.~Campana \emph{et~al.}, ``Pdf--ecg in
  clinical practice: A model for long--term preservation of digital 12--lead
  ecg data,'' \emph{Journal of electrocardiology}, vol.~50, no.~6, pp.
  776--780, 2017.

\bibitem{wagner2014marriott}
G.~S. Wagner and D.~G. Strauss, \emph{Marriott's practical
  electrocardiography}.\hskip 1em plus 0.5em minus 0.4em\relax Lippincott
  Williams \& Wilkins, 2014.

\bibitem{kligfield2007recommendations}
P.~Kligfield, L.~S. Gettes, J.~J. Bailey, R.~Childers, B.~J. Deal, E.~W.
  Hancock, G.~Van~Herpen, J.~A. Kors, P.~Macfarlane, D.~M. Mirvis
  \emph{et~al.}, ``Recommendations for the standardization and interpretation
  of the electrocardiogram: part i: the electrocardiogram and its technology a
  scientific statement from the american heart association electrocardiography
  and arrhythmias committee, council on clinical cardiology; the american
  college of cardiology foundation; and the heart rhythm society endorsed by
  the international society for computerized electrocardiology,'' \emph{Journal
  of the American College of Cardiology}, vol.~49, no.~10, pp. 1109--1127,
  2007.

\bibitem{eberz2017broken}
S.~Eberz, N.~Paoletti, M.~Roeschlin, M.~Kwiatkowska, I.~Martinovic, and
  A.~Patan{\'e}, ``Broken hearted: How to attack ecg biometrics,'' in
  \emph{Network and Distributed System Security Symposium 2017}.\hskip 1em plus
  0.5em minus 0.4em\relax Internet Society, 2017.

\bibitem{Wu2021}
S.-C. Wu, P.-L. Hung, and A.~L. Swindlehurst, ``Ecg biometric recognition:
  Unlinkability, irreversibility, and security,'' \emph{IEEE Internet of Things
  Journal}, vol.~8, no.~1, pp. 487--500, 2021.

\bibitem{awadallah2024artificial}
A.~Awadallah, K.~Eledlebi, J.~Zemerly, D.~Puthal, E.~Damiani, K.~Taha, T.-Y.
  Kim, P.~D. Yoo, K.-K.~R. Choo, M.-S. Yim \emph{et~al.}, ``Artificial
  intelligence-based cybersecurity for the metaverse: research challenges and
  opportunities,'' \emph{IEEE Communications Surveys \& Tutorials}, 2024.

\bibitem{shandhi2020estimation}
M.~M.~H. Shandhi, W.~H. Bartlett, J.~A. Heller, M.~Etemadi, A.~Young,
  T.~Pl{\"o}tz, and O.~T. Inan, ``Estimation of instantaneous oxygen uptake
  during exercise and daily activities using a wearable
  cardio-electromechanical and environmental sensor,'' \emph{IEEE journal of
  biomedical and health informatics}, vol.~25, no.~3, pp. 634--646, 2020.

\bibitem{ji2024ecg}
W.~Ji and D.~Zhu, ``Ecg classification exercise health analysis algorithm based
  on gru and convolutional neural network,'' \emph{IEEE Access}, vol.~12, pp.
  59\,842--59\,850, 2024.

\bibitem{tasnim2023cardiovascular}
M.~Tasnim, A.~J. Patinga, H.~Shahriar, and S.~Sneha, ``Cardiovascular health
  management compliance with health insurance portability and accountability
  act,'' in \emph{2023 IEEE 47th Annual Computers, Software, and Applications
  Conference (COMPSAC)}.\hskip 1em plus 0.5em minus 0.4em\relax IEEE, 2023, pp.
  1423--1428.

\bibitem{9427539}
A.~N. Navaz, M.~A. Serhani, H.~T. El~Kassabi, N.~Al-Qirim, and H.~Ismail,
  ``Trends, technologies, and key challenges in smart and connected
  healthcare,'' \emph{IEEE Access}, vol.~9, pp. 74\,044--74\,067, 2021.

\bibitem{24}
Z.~Qu, W.~Shi, B.~Liu, D.~Gupta, and P.~Tiwari, ``Iomt-based smart healthcare
  detection system driven by quantum blockchain and quantum neural network,''
  \emph{IEEE journal of biomedical and health informatics}, vol.~28, no.~6, pp.
  3317--3328, 2023.

\bibitem{25}
W.~Yang and S.~Wang, ``A privacy-preserving ecg-based authentication system for
  securing wireless body sensor networks,'' \emph{IEEE Internet of Things
  Journal}, vol.~9, no.~8, pp. 6148--6158, 2021.

\bibitem{26}
B.~Samuel and M.~K. Hota, ``Dual attention based pipelined encoder-decoder
  network for fetal electrocardiogram extraction,'' \emph{IEEE Transactions on
  Instrumentation and Measurement}, 2025.

\bibitem{27}
G.~Wang, S.~Shanker, A.~Nag, Y.~Lian, and D.~John, ``Ecg biometric
  authentication using self-supervised learning for iot edge sensors,''
  \emph{IEEE Journal of Biomedical and Health Informatics}, 2024.

\bibitem{28}
C.~Tan, L.~Zhang, T.~Qian, S.~Br{\'a}s, and A.~J. Pinho, ``Statistical n-best
  afd-based sparse representation for ecg biometric identification,''
  \emph{IEEE Transactions on Instrumentation and Measurement}, vol.~70, pp.
  1--13, 2021.

\bibitem{28b}
T.~Qian, L.~Zhang, and Z.~Li, ``Algorithm of adaptive fourier decomposition,''
  \emph{IEEE Transactions on Signal Processing}, vol.~59, no.~12, pp.
  5899--5906, 2011.

\bibitem{29}
P.~Huang, L.~Guo, M.~Li, and Y.~Fang, ``Practical privacy-preserving ecg-based
  authentication for iot-based healthcare,'' \emph{IEEE Internet of Things
  Journal}, vol.~6, no.~5, pp. 9200--9210, 2019.

\bibitem{30}
M.~Usman, M.~Kamal, and M.~Tariq, ``Improved and secured electromyography in
  the internet of health things,'' \emph{IEEE journal of biomedical and health
  informatics}, vol.~26, no.~5, pp. 2032--2040, 2021.

\bibitem{31}
W.~Dargie, S.~Farrokhi, and C.~Poellabauer, ``Identification of persons based
  on electrocardiogram and motion data,'' \emph{IEEE Sensors Journal}, 2025.

\end{thebibliography}


\end{document}